\documentclass{article}

\usepackage{arxiv}

\usepackage[utf8]{inputenc}
\usepackage[T1]{fontenc}
\usepackage{hyperref}
\usepackage{url}
\usepackage{booktabs}
\usepackage{amsfonts}
\usepackage{amsmath}
\usepackage{nicefrac}
\usepackage{microtype}
\usepackage{cleveref}
\usepackage{graphicx}
\usepackage{natbib}
\usepackage{doi}
\usepackage{multirow}
\usepackage{array}

\title{Social World Model for Lifelong Social Intelligence}

\date{}

\author{
  Yu Luo \\
  Central South University \\
}

\renewcommand{\shorttitle}{Social World Model for Lifelong Social Intelligence}

\hypersetup{
  pdftitle={Social World Model for Lifelong Social Intelligence},
  pdfsubject={cs.AI, cs.CL},
  pdfauthor={Anonymous},
  pdfkeywords={social intelligence, lifelong learning, world model, preference optimization, language agents},
}

\begin{document}
\maketitle

\begin{abstract}
Social intelligence has emerged as a core competency for language agents operating in real-world collaborative scenarios.
Recent advances in open-ended social evaluation and multi-turn lifelong assessment have significantly sharpened our ability to measure social capabilities and have begun to characterize their performance boundaries in complex interactions.
However, current evidence remains largely confined to ``capability characterization and comparison,'' with insufficient mechanistic explanation of how such capabilities are continuously shaped and stably accumulated.
This gap calls for a shift from static evaluation paradigms toward sustainable learning paradigms in social intelligence research.

Two critical pain points emerge at the methodological level.
First, social interaction trajectories lack a unified structured state representation and traceable data constraints, making it difficult for experience to crystallize into retrainable, auditable, and iterable learning signals.
Second, capability improvement and capability retention have been studied under separate experimental settings, lacking joint measurement and causal attribution under a single protocol.
As a result, research more readily yields static observations of short-term gains, but struggles to establish testable evidence regarding the continuous evolution of capabilities.

To bridge this gap, we propose the \textbf{Social World Model}.
We decompose social interaction into five dimensions---scene setting, observation, mental state, action, and dialogue---and build a closed-loop learning framework upon this decomposition: the agent collects experience during interaction, converts trajectories into preference signals for model updating, and redeploys the updated policy into the next round of social environments for continued learning.
Simultaneously, we provide a reusable data synthesis mechanism for social lifelong learning, together with a social lifelong learning benchmark that enables subsequent researchers to train and directly evaluate on the same data.
Compared to evaluation-centric existing frameworks, our approach further transforms the ``object of evaluation'' into an ``object of sustainable training.''

To validate the framework, we conduct experiments on the social lifelong learning benchmark ASCENT-Bench.
After interactive training, the Qwen2.5-7B model outperforms its untrained baseline across all five core metrics.
Moreover, the trained model matches the closed-source Gemini~3 Flash on completion rate, exceeds Gemini on pass rate, and achieves zero forgetting across three difficulty levels.
In contrast to prior work that only reports social capability decay or static comparisons, this paper provides an end-to-end pathway that is ``trainable, verifiable, and retainable,'' demonstrating that small open-source models can acquire competitive and sustainable social coordination capabilities through closed-loop social world training.
\end{abstract}


\section{Introduction}

\begin{figure}[htbp] 
    \centering 
    \includegraphics[width=1\textwidth]{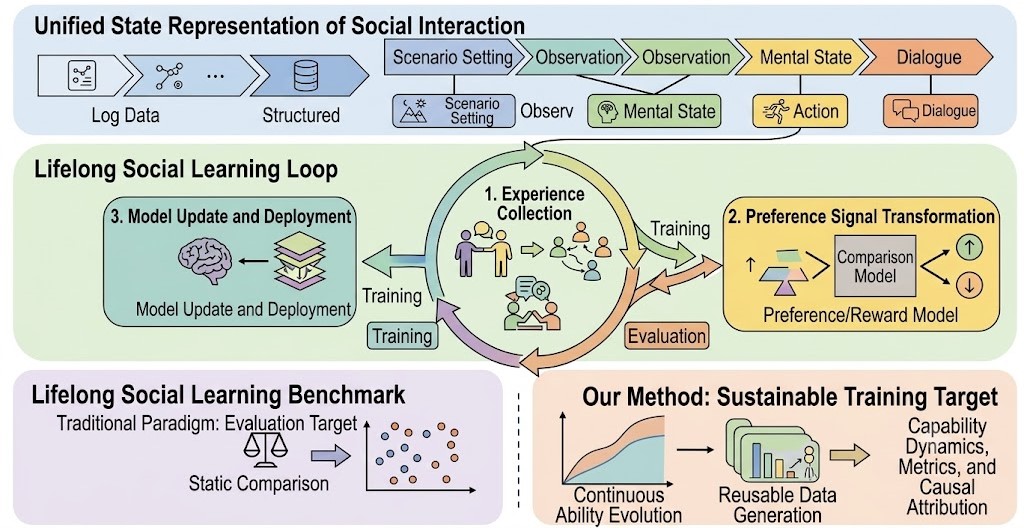}
    \caption{Overall} 
    \label{fig:overall} 
\end{figure}

Social intelligence means more than generating polite, natural conversational text.
A truly socially capable agent must be able to infer others' goals and preferences, act under incomplete information, negotiate among conflicting constraints, maintain cooperative relationships while advancing its own task objectives, and adjust subsequent behavior based on prior interaction outcomes.
Current language agents already exhibit strong surface-level social expression, yet mainstream training and evaluation paradigms remain static: models are trained once, evaluated on fixed prompts or short-horizon interactions, and then assigned a ``social capability level.''
This approach neglects one of the most essential properties of social intelligence---that it is fundamentally a capability that improves continuously through action, feedback, correction, and redeployment in sustained interaction.

Prior work has made significant progress in evaluating social capability and mental state reasoning.
SOTOPIA formulates social intelligence as open-ended role-playing oriented toward social goals, demonstrating that social capability exhibits multi-agent, goal-directed, and scenario-dependent characteristics~\citep{zhou2023sotopia}.
ToMBench and HI-TOM respectively approach the problem from systematic theory-of-mind tasks and higher-order recursive mental state reasoning, showing that ``understanding what others think'' must be modeled as a decomposable, testable, and comparable core capability~\citep{chen2024tombench,he2023hitom}.
Generative Agents further demonstrate that when LLM agents possess memory, reflection, planning, and interaction mechanisms, they can exhibit sustained and credible social behavior in sandbox environments~\citep{park2023generative}.
These works establish that social behavior can be simulated and measured.
The critical step this paper advances is unifying ``simulation,'' ``training,'' and ``evaluation'' into a single continuously iterable closed loop: continuously collecting social experience, converting it into high-quality training signals, using them to update smaller open-source models, and verifying whether such updates truly bring stable social capability improvements.

We argue that social intelligence should be understood as a lifelong world modeling problem.
A socially intelligent agent must form internal representations of its interaction environment: who the participants are, what scene constraints hold, what beliefs and preferences each party may hold, what action sets are available, and how dialogue changes subsequent states.
Building on this insight, we combine classical world model thinking~\citep{ha2018world} with the wake-sleep learning mechanism~\citep{hinton1995wake} in the social intelligence learning process: the \emph{wake} phase collects interaction experience, the \emph{sleep} phase converts experience into learnable supervision and completes parameter updates, and the \emph{deploy} phase places the updated policy back into the social environment for testing and accumulation.
Unlike one-time static fine-tuning, this process emphasizes continuous correction and long-term improvement through interaction consequences.

This paper makes three main contributions.
First, we propose the Social World Model framework for lifelong social intelligence, explicitly decomposing social interaction into five dimensions---scene setting, observation, mental state, action, and dialogue---so that cross-episode experience can be systematically recorded, tracked, and reused.
Second, we establish a data synthesis and unified evaluation system for lifelong learning.
The benchmark covers eight types of social coordination tasks, three difficulty tiers, and a three-cycle progressive protocol, reporting not only average performance but also jointly characterizing forward transfer, backward forgetting, adaptive difficulty, and hierarchical pass-rate evolution.
Third, we provide empirical evidence of interaction-driven learning: a 7B open-source model trained with GRPO cycles outperforms its untrained baseline across all five core metrics, matches the Gemini proxy model on completion rate, and exceeds Gemini on pass rate.

Thus, our core research question is: \emph{Can a small open-source model improve its social coordination capability through a closed-loop social world model, and can this improvement be reliably measured under a benchmark that jointly accounts for capability gain and lifelong stability?}
Under the ASCENT-Bench protocol, our results provide an affirmative answer, with evidence most pronounced in task completion rate, pass rate, action efficiency, and resistance to forgetting.

\section{Related Work}

\begin{table}[t]
\caption{Comparison with representative frameworks across key dimensions.}
\label{tab:related}
\centering
\small
\begin{tabular}{lccccc}
\toprule
\textbf{Dimension} & \textbf{SOTOPIA} & \textbf{Lifelong SOTOPIA} & \textbf{Gen. Agents} & \textbf{Voyager} & \textbf{Ours} \\
\midrule
Training mechanism     & None          & None          & None             & Skill library    & GRPO + dual-track \\
Data generation        & Human eval    & Human eval    & Memory-reflection & Auto curriculum  & Structured 5-dim traj. \\
Evaluation protocol    & Single-shot   & Multi-cycle   & Behavioral cred. & Skill completion & Progressive + forget. \\
Forgetting detection   & No            & Yes (backward)& No               & No               & Cross-tier zero-forget \\
Forward transfer       & No            & Yes           & No               & Open exploration & Hierarchical pass rate \\
Preference signal      & Human rating  & Human rating  & No               & No               & Auto multi-dim + audit \\
Closed-loop iteration  & No            & Eval loop     & No               & No               & Train-eval-deploy loop \\
\bottomrule
\end{tabular}
\end{table}

\subsection{Social Intelligence and Theory-of-Mind Benchmarks}

SOTOPIA constructs social intelligence as interactive role-playing centered on social goals and evaluates models through a holistic social interaction protocol~\citep{zhou2023sotopia}.
This is directly inspiring for our work, as it moves beyond isolated question-answering and treats social capability as an emergent outcome of multi-turn interaction.
However, SOTOPIA primarily evaluates what capabilities a model \emph{currently} possesses, whereas we focus on the \emph{lifelong} dimension: how the same model family exhibits measurable capability changes across progressive difficulty and forgetting checks before and after social world training.

Theory-of-mind benchmarks provide another critical foundation.
ToMBench constructs a systematic inventory of theory-of-mind tasks, while HI-TOM focuses on higher-order recursive mental state reasoning~\citep{chen2024tombench,he2023hitom}.
Together, these works demonstrate that the key to social intelligence lies not in surface phrasing but in structured modeling of others' beliefs, preferences, constraints, and interaction consequences.
We build on this foundation by explicitly using mental states and diagnostic signals in our system, and by separately measuring understanding, execution, subjective experience, comprehensive social diagnosis, and objective task completion through metrics such as $U_{\text{final}}$, $A_{\text{eff}}$, $\text{NPC}_{\text{final}}$, $S_{\text{diag}}$, and $\text{CompletionRate}$.

Lifelong SOTOPIA is closest to our work in research objectives, as it explicitly studies social intelligence in lifelong social interaction~\citep{goel2025lifelong}.
However, two key differences remain.
First, we provide not only an evaluation perspective but a complete training-evaluation closed loop: trajectories are converted into preference pairs, the action track and dialogue track are trained separately, and deployable adapters are produced.
Second, our benchmark explicitly introduces progressive difficulty, forgetting metrics, and auditable observation diagnostics, making it closer to a continual learning protocol than a one-shot social test.

\subsection{Interactive Agents and Social Simulation}

Generative Agents demonstrated that when LLM agents possess memory, reflection, and planning capabilities, they can simulate credible everyday social behavior~\citep{park2023generative}.
The significance of this work lies in showing that an interaction environment is not merely a display layer but can serve as a structured source of behavioral trajectories.
Our Social World Model inherits this idea but changes the objective: we care not only about whether agents ``look credible'' in a sandbox, but whether the interaction world can stably produce auditable training data and genuinely improve the deployed model's capabilities in subsequent cycles.

Voyager demonstrated a related approach in the Minecraft setting: an LLM-based embodied agent can accumulate skills through automatic curricula and skill libraries, continuously improving exploration in open environments~\citep{wang2023voyager}.
However, social tasks differ fundamentally.
Social tasks cannot be solved merely by reusing physical skills; they require the agent to coordinate preferences across multiple NPCs, handle conflicts, maintain relationships, and track hidden or partially observable mental states.
Therefore, we need a benchmark and data pipeline that treats ``social consequences'' as a first-class training signal.

\subsection{Continual Learning}

Continual learning concerns how models can acquire new capabilities without forgetting existing ones.
Parisi et al.\ systematically review this field, noting that sequentially trained systems are inherently susceptible to catastrophic forgetting unless mechanisms for retaining or replaying prior knowledge are introduced~\citep{parisi2019continual}.
French's classic analysis of catastrophic forgetting further reveals the intrinsic reasons why sequential updates overwrite existing behavioral patterns~\citep{french1999catastrophic}.
These insights directly shape our benchmark design: we do not treat improvement on new tasks as a sufficient condition, but explicitly measure backward forgetting across tiers and report that our trained model achieves zero forgetting across all three difficulty levels.

\subsection{World Models}

World models provide the theoretical foundation for ``learning from internally simulated experience.''
Ha and Schmidhuber propose learning compact representations of environment dynamics to support agent behavior~\citep{ha2018world}.
This idea has been widely adopted in visual control and planning tasks, but exploration in social interaction scenarios remains limited.
The key distinction between social and physical environments is that state transitions depend not only on the agent's own actions but also on the beliefs, preferences, and strategies of other participants---information that is often partially observable or entirely invisible.
Therefore, migrating world model thinking to the social domain requires an extended framework capable of explicitly representing mental states and relational dynamics.

\subsection{Wake-Sleep Learning}

Hinton et al.'s wake-sleep algorithm distinguishes between recognition updates driven by real data and sleep updates driven by the generative phase~\citep{hinton1995wake}.
We adapt this alternating learning mechanism to the lifelong learning process of social language agents: the wake phase collects social interaction trajectories, the sleep phase converts trajectories into learnable supervision and completes parameter updates, and the deploy phase places the updated policy back into the social environment for testing and accumulation.
Unlike traditional single-pass fine-tuning, this closed-loop structure enables the model to continuously correct and improve from the consequences of its own interactive behavior.

\subsection{Preference Optimization for Language Models}

Instruction following and preference optimization have become standard tools for language model alignment.
InstructGPT improves model adherence to user intent through supervised fine-tuning, reward modeling, and reinforcement learning from human feedback~\citep{ouyang2022instructgpt}.
Direct Preference Optimization (DPO) demonstrates that one can directly optimize policy from pairwise preferences without training an explicit reward model~\citep{rafailov2023dpo}.
DeepSeekMath introduces GRPO as an efficient group-relative policy optimization method and proves its effectiveness in reasoning task training~\citep{shao2024deepseekmath}.

We explore the application of GRPO in interactive social coordination tasks.
In this setting, the reward objective is modeled as a multi-dimensional social feedback signal composed of understanding, action effectiveness, NPC satisfaction, diagnostic score, and completion.
Thus, we combine preference optimization with fine-grained metric source control and action/dialogue dual-track collaborative training.
The methodological core of this paper is that only when GRPO is placed within a social environment closed loop that produces structured, source-auditable multi-dimensional feedback signals can preference optimization be truly transformed into improvements in model social capability.

\section{Social World Model}

\subsection{Problem Definition}

We define a social world as a partially observable multi-agent environment.
In each episode, a target agent interacts with several NPCs around a specific scene goal.
The system state includes the task setup, visible observations, latent or estimated mental states, historical actions, dialogue history, and constraints such as budget, time, preferences, or social pressure.
The agent must select actions and utterances that advance the goal while maintaining acceptable social outcomes.

A social trajectory can be represented as a structured record sequence:
\begin{equation}
  \tau = \{(s_t, o_t, m_t, a_t, d_t, r_t)\}_{t=1}^{T},
\end{equation}
where $s_t$ denotes the scene state, $o_t$ the observation, $m_t$ the mental state estimate, $a_t$ the selected action, $d_t$ the dialogue output, and $r_t$ the multi-dimensional evaluation signal.
In our implementation, the evaluation signal includes understanding, application efficiency, NPC satisfaction, diagnostic score, and a value derived from completion.
This representation transforms social interaction into trainable data rather than unstructured chat logs.

\subsection{Five-Dimensional Interaction Decomposition}

The Social World Model decomposes each social episode into five dimensions.

\paragraph{Setting} defines the task context, including participants, goals, resources, constraints, and conflict structures.
In our benchmark, scenarios include team lunch coordination, book club resource trade-offs, birthday party budget conflicts, charity fundraising, community market organization, hackathon coordination, weather-risk picnic, and activity debrief.

\paragraph{Observation} records the information currently perceivable by the agent: existing dialogue, visible participants, current sub-goal status, and available actions.
This ensures that observational evidence enters the auditable training and evaluation chain directly.

\paragraph{Mental State} records estimates of NPC preferences, beliefs, emotions, and social constraints.
This dimension is critical because social success often depends on inferring unstated preferences rather than merely responding to explicit instructions.

\paragraph{Action} records strategic actions: whom to approach, which sub-goal to prioritize, whether to negotiate, wait, inquire, propose, or repair conflicts.
Action quality is evaluated separately from surface dialogue quality.

\paragraph{Dialogue} records the natural language realization of actions.
This separation allows the training system to distinguish between two distinct error types: ``the strategy itself was correct, but the expression was poor'' and ``the expression was polite, but the task was not advanced.''

\subsection{Wake-Sleep-Deploy Closed-Loop Learning}

The framework follows a wake-sleep-deploy closed loop.
In the \emph{wake} phase, the current model interacts with the social environment and generates trajectories.
In the \emph{sleep} phase, the system converts trajectories into preference data, completes parameter updates after quality screening, and produces auditable training records.
In the \emph{deploy} phase, the updated model is re-tested in the social environment, thereby initiating the next learning cycle.

From a data processing perspective, the entire closed loop comprises the following stages: raw trajectory loading and structuring, long-trajectory segmentation and counterfactual candidate generation, step-level and trajectory-level preference pair construction, and policy optimization based on preference pairs.
Each stage produces traceable intermediate artifacts, ensuring that every step from raw interaction experience to final parameter update is auditable and reproducible.

During trajectory processing, the system first structures raw interaction records into a unified format and segments long trajectories appropriately.
Subsequently, the system generates counterfactual candidate actions and their expected return estimates for each decision step, used to construct step-level preference pairs.
Simultaneously, the system extracts trajectory-level preference evidence from complete trajectories to construct dialogue-level preference pairs.
This multi-level preference construction mechanism ensures that training data consists not of loose dialogue collections but of social decision records strictly bound to interaction outcomes.

\subsection{Dual-Track Action and Dialogue Optimization}

Social capability exhibits a characteristic multi-objective nature: strategy selection and language expression are both interdependent and potentially contradictory.
To address this, the training process separates action and dialogue into two independent optimization tracks.
The \emph{action track} is based on step-level preference pairs, focusing on decision quality, task advancement, and execution efficiency; the \emph{dialogue track} is based on trajectory-level preference pairs, focusing on language expression appropriateness and relationship maintenance.
Each track has its own data screening, training, and evaluation pipeline, thereby preserving learning evidence from different sources.

The necessity of this separation lies in the fact that a model may select the correct strategy but express it in a way that damages cooperative relationships, or it may be linguistically polite yet fail to advance the task.
Using a single undifferentiated reward signal often fails to reveal this tension.
By training separately, we can independently examine the source of performance improvement---whether it stems from enhanced strategy selection, improved language expression quality, or both.

Before training, both tracks undergo a unified data screening pipeline, including model capability filtering, duplicate template suppression, sample deduplication, and quality threshold screening.
Duplicate template suppression prevents the model from learning from highly homogeneous interaction patterns; quality threshold screening ensures that low-signal samples do not dominate parameter updates.
Additionally, the system supports supplementing data from historical cycles to alleviate insufficient data volume in a single cycle.

\subsection{Strict Metric Gating Mechanism}

Strict metric gating is the most critical data quality assurance mechanism in the entire framework.
Its core objective is to prevent the model from learning from spurious metrics or low-confidence evaluation signals.
In our experimental setup, the action track mandates that every training sample be accompanied by valid multi-dimensional metric evidence; the dialogue track allows moderate relaxation of acceptance criteria when structured metric density is insufficient, to maintain training data scale.

Under strict gating, each training sample must contain valid values for four metrics: understanding, action efficiency, NPC satisfaction, and comprehensive diagnosis.
These metrics must pass not only numerical validity checks but also source traceability checks: if a metric source label is missing, derived from proxy inference, or flagged as a degraded estimate, the sample is directly excluded.
During reward computation, if the model output cannot be parsed into structured metrics, the system assigns a penalty reward rather than relying on heuristic estimates.
This design is particularly important because social training is highly susceptible to reward hacking---the model may generate seemingly appropriate social expressions that neither complete the task nor maintain reliable mental state reasoning.

Furthermore, the system preserves source annotation at the observation level, distinguishing between directly structured evidence and proxy-inferred evidence.
Low-confidence observations are flagged as proxy inferences and do not enter the strict metric main chain.
This mechanism ensures that core benchmark conclusions are driven by verifiable metrics while maintaining traceability of diagnostic information.

\section{Social Lifelong Learning Benchmark}

\begin{figure}[htbp] 
    \centering 
    \includegraphics[width=1\textwidth]{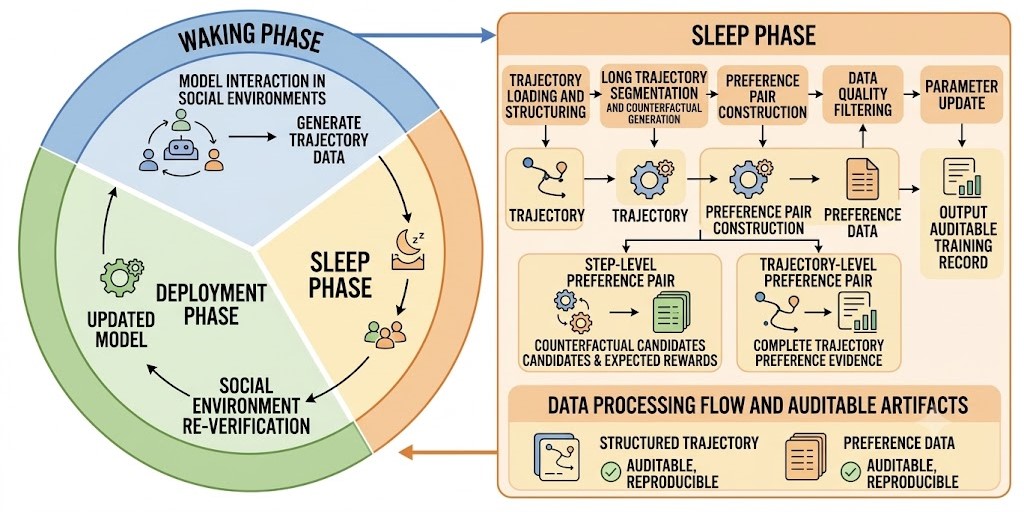}
    \caption{Experiment} 
    \label{fig:experiment} 
\end{figure}

\subsection{Benchmark Arms and Scenarios}

The benchmark evaluates three model types under a unified set of eight social coordination scenarios: an untrained 7B open-source baseline (Arm~A), the same-scale open-source model after closed-loop training cycles (Arm~B), and a high-performance closed-source proxy model (Arm~C).
This design aims to answer two core questions: whether closed-loop training yields stable gains (comparing A and B), and whether such gains can push the small model to a level comparable to a strong proxy (comparing B and C).

The eight scenarios are organized into three difficulty tiers.
L0 scenarios have clear constraints and relatively simple preference structures; L1 scenarios introduce multi-objective coordination and more complex resource-relationship trade-offs; L2 scenarios further increase dynamics, conflict intensity, and negotiation pressure.
Through hierarchical design, the benchmark distinguishes between capabilities that are ``effective in simple situations'' and those that ``remain effective under high-pressure complex situations.''

\subsection{Progressive Unlocking and Lifelong Evaluation}

The benchmark employs a three-cycle progressive unlocking protocol.
Cycle~1 evaluates only L0; Cycle~2 expands to L0+L1; Cycle~3 covers L0+L1+L2.
In each cycle, each available scenario runs 3 episodes.
Thus, each model arm comprises 45 episodes, totaling 135 episodes across three arms.

This design evaluates not single-shot static performance but the dynamic evolution of capabilities as task complexity increases.
If a model benefits only on low-difficulty tasks, its advantage may not transfer to subsequent tiers; if a model trades existing capabilities for new ones, observable forgetting will appear at earlier tiers.
Accordingly, the benchmark simultaneously reports hierarchical pass-rate evolution, forward transfer, backward forgetting, and adaptive difficulty behavior.

\subsection{Core Metrics}

The benchmark includes five core metrics.

\paragraph{$U_{\text{final}}$} measures understanding capability.
When step-level alignment scores are available, it is defined as $0.5 \times \text{mean}(\text{alignmentScores}) + 0.5 \times \text{completionRate}$; otherwise it degrades to $\text{completionRate}$.
This design anchors understanding evaluation to task completion, reducing the risk of subjective self-assessment signals dominating conclusions.

\paragraph{$A_{\text{eff}}$} measures application efficiency.
When step-level rewards are available, it is defined as $0.5 \times \text{mean}(\text{stepRewards}) + 0.5 \times (\text{completionRate} \times 2 - 1)$; otherwise it is derived from $\text{completionRate}$.
This definition directly links efficiency evaluation to objective task advancement.

\paragraph{$\text{NPC}_{\text{final}}$} measures NPC satisfaction on a 1--5 Likert scale.
When step-level satisfaction is available, it is fused with completion-derived satisfaction to mitigate single-source noise.

\paragraph{$S_{\text{diag}}$} is the comprehensive social diagnostic score, defined as:
\begin{equation}
  S_{\text{diag}} = 0.4 \times U_{\text{final}} + 0.4 \times \frac{A_{\text{eff}} + 1}{2} + 0.2 \times \frac{\text{NPC}_{\text{final}} - 1}{4}.
\end{equation}
This weighting places ``understanding'' and ``action'' on equal footing while preserving the relationship quality dimension.

\paragraph{CompletionRate} is the proportion of completed sub-goals, one of the most objective outcome metrics since it is computed directly from goal states.
The benchmark's pass criterion is: an episode is recorded as passed when $\text{completionRate} \geq 0.5$ and $S_{\text{diag}} \geq 0.45$.

\subsection{Diagnostic Layer and Auditable Observation}

The diagnostic layer provides retrospective evidence parallel to the core scoring.
For each episode, the system extracts representative steps and generates observation summaries while recording behavioral consistency signals to determine whether strategy selection aligns with observable facts, such as whether the target object is interactable, whether actions point to valid sub-goals, and whether avoidable inefficient waiting exists.

In the current benchmark run, the diagnostic layer covers 135/135 episodes with an average confidence of 0.98 and is independent of the core scoring system.
Its role is to provide traceable behavioral consistency evidence for each episode, supporting structured attribution of failure sources, such as goal identification bias, inappropriate conflict handling sequence, inefficient interaction, and strategy-expression mismatch.

For social world training, such evidence directly enhances the interpretability of training signals: the system judges not only ``whether passed'' but also ``why passed or failed,'' and accordingly improves subsequent data screening and preference construction, forming a more targeted optimization loop.

\section{Experiments and Results}

\subsection{Experimental Setup}

The experiment compares Arms~A, B, and C on the ASCENT-Bench.
All three use the identical eight scenarios, three-cycle progressive unlocking schedule, and the same metric definitions.
Each model arm comprises 45 episodes, totaling 135 episodes.
The primary research questions decompose into two comparisons: A vs.\ B to verify whether social world training improves the open-source 7B model; B vs.\ C to test whether the trained 7B model approaches a stronger closed-source proxy.

The trained B model is obtained from the aforementioned closed-loop learning framework.
This framework combines GRPO preference optimization, action/dialogue dual-track learning, and strict metric gating during training to ensure that parameter updates are built on traceable and high-confidence social evidence.
Training summaries, preference samples, and model version records are retained throughout the experiment to support result verification.

\subsection{Core Results: A $<$ B}

\begin{table}[t]
\caption{Core comparison between untrained baseline (A) and trained model (B). B outperforms A across all five core metrics.}
\label{tab:ab}
\centering
\begin{tabular}{lccc}
\toprule
\textbf{Metric} & \textbf{A: Untrained 7B} & \textbf{B: Trained 7B} & \textbf{B/A} \\
\midrule
$U_{\text{final}}$       & 0.584  & 0.651  & 1.11$\times$ \\
$A_{\text{eff}}$         & 0.134  & 0.264  & 1.96$\times$ \\
$S_{\text{diag}}$        & 0.576  & 0.641  & 1.11$\times$ \\
CompletionRate            & 0.436  & 0.582  & 1.33$\times$ \\
PassRate                  & 37.8\% & 64.4\% & 1.70$\times$ \\
\bottomrule
\end{tabular}
\end{table}

Table~\ref{tab:ab} presents the core result: B outperforms A on all five core metrics.
The largest relative improvement is in $A_{\text{eff}}$, increasing from 0.134 to 0.264, nearly doubling.
Meanwhile, CompletionRate increases from 0.436 to 0.582, and PassRate from 37.8\% to 64.4\%.
Since the pass rate jointly constrains task completion and comprehensive social diagnostic quality, its improvement better reflects genuine enhancement in social coordination capability.

$U_{\text{final}}$ and $S_{\text{diag}}$ show more modest but consistently positive improvements, with absolute gains of 0.067 and 0.065 respectively.
This indicates that model improvement is not confined to a single efficiency dimension but exhibits simultaneous positive changes in understanding capability and comprehensive social diagnostic capability.

\subsection{Comparative Experiment: Full Framework vs.\ Alternative Training Paths}

To answer the stricter question of ``where the performance gains actually come from,'' we further compare the full framework M4 against four executable control paths: untrained baseline M0, supervised fine-tuning only M1, vanilla GRPO only M2, and AMPO-style DPO path M3.
All methods are evaluated under the same ASCENT-Bench protocol, using the identical scenario pool, three-cycle progressive unlocking setup, and 45-episode budget.
Final auditing confirms that all five groups yield valid completed experiments.

\begin{table}[t]
\caption{Comparison across methods. M0: untrained baseline; M1: SFT only; M2: vanilla GRPO; M3: AMPO-style DPO; M4: full Social World framework.}
\label{tab:methods}
\centering
\small
\begin{tabular}{lccccc}
\toprule
\textbf{Method} & $U_{\text{final}}$ & $A_{\text{eff}}$ & $S_{\text{diag}}$ & \textbf{CR} & \textbf{PR} \\
\midrule
M0: Untrained baseline   & 0.584 & 0.134 & 0.576 & 0.436 & 37.8\% \\
M1: SFT only             & 0.450 & 0.012 & 0.484 & 0.444 & 48.9\% \\
M2: Vanilla GRPO         & 0.310 & $-$0.174 & 0.370 & 0.338 & 22.2\% \\
M3: AMPO-style DPO       & 0.072 & $-$0.629 & 0.140 & 0.000 & 0.0\% \\
M4: Full framework       & 0.651 & 0.264 & 0.641 & 0.582 & 64.4\% \\
\bottomrule
\end{tabular}
\end{table}

These results first provide a clear failure case: it is not the case that ``applying preference optimization to social trajectories automatically improves capability.''
M2's vanilla GRPO and M3's AMPO-style DPO both fail to produce stable gains, instead showing significant degradation on key outcome metrics; M3's completion rate and pass rate both drop to 0.
This means the difficulty of social learning lies not in whether modern optimization methods are used, but in whether these methods can receive training signals consistent with genuine coordination outcomes.

In contrast, the full framework M4 achieves the best results across all five core metrics, outperforming both the untrained M0 and the three ``partial method'' or ``alternative optimization path'' variants.
Compared to M1, M4 provides additional improvements of 0.201, 0.252, 0.158, 0.138, and 15.5 percentage points on $U_{\text{final}}$, $A_{\text{eff}}$, $S_{\text{diag}}$, CompletionRate, and PassRate respectively, indicating that static supervised fine-tuning alone cannot learn strategy execution and relationship maintenance in social coordination.
Compared to M2, M4 exceeds by 42.2 percentage points on pass rate and turns $A_{\text{eff}}$ from negative to positive, further demonstrating that without the structured evidence constraints of our framework, GRPO alone does not automatically converge to better social behavior.

These control experiments thus support not a generic ``full system outperforms simplified system'' conclusion, but a more specific mechanistic judgment: the continuous shaping of social capability depends on strict alignment among training signals, capability objectives, and evaluation protocols.
Five-dimensional world modeling transforms social interaction into sustainably exploitable training objects; dual-track preference optimization distinguishes strategy and expression evidence; and the wake-sleep-deploy closed loop ensures that the model must be accountable for its update results in the next interaction round.
Only when these three are unified under a single protocol do performance improvements emerge stably.

Beyond internal method comparisons, we also supplement two external comparison experiments to verify result transferability.
First, the AgentSense formal-45 comparison has been completed, with both arms producing complete outputs on a fixed 45-sample subset, demonstrating that our framework supports cross-benchmark, cross-script executable verification.
Second, the SOTOPIA formal-45 comparison has also been completed, with both arms successfully running on 45 social scenarios, forming external interaction evidence independent of ASCENT-Bench.
It should be emphasized that these two experiments do not share exactly the same metric definitions and judges as the main benchmark, so their role is to provide external consistency support rather than directly replace the primary result ranking on ASCENT-Bench.

\subsection{Comparison with the Gemini Proxy}

\begin{table}[t]
\caption{Comparison between trained 7B model (B) and Gemini proxy (C). B matches C on completion rate and exceeds C on pass rate.}
\label{tab:bc}
\centering
\begin{tabular}{lccc}
\toprule
\textbf{Metric} & \textbf{B: Trained 7B} & \textbf{C: Gemini Proxy} & \textbf{B/C} \\
\midrule
$U_{\text{final}}$       & 0.651  & 0.748  & 87\%  \\
$A_{\text{eff}}$         & 0.264  & 0.400  & 66\%  \\
$S_{\text{diag}}$        & 0.641  & 0.715  & 90\%  \\
CompletionRate            & 0.582  & 0.582  & 100\% \\
PassRate                  & 64.4\% & 60.0\% & 107\% \\
\bottomrule
\end{tabular}
\end{table}

Table~\ref{tab:bc} reports the comparison between B and C.
B does not outperform Gemini on all dimensions, but under the 20\% proximity criterion adopted in this paper, it reaches levels close to or exceeding Gemini on four out of five metrics.

The trained 7B model exactly matches Gemini on completion rate and exceeds Gemini on pass rate; meanwhile, $U_{\text{final}}$ and $S_{\text{diag}}$ reach 87\% and 90\% of Gemini's values respectively, demonstrating stable high proximity.
This result indicates that social world training can significantly enhance outcome-level social coordination capability while preserving the scale advantage of open-source small models, achieving competitive performance at the same magnitude as strong closed-source proxies on multiple core metrics.

This result supports a key conclusion: social world training significantly narrows the performance gap between open-source small models and strong closed-source proxies on structured social coordination tasks.
Multi-metric evidence shows that our method not only improves outcome-level performance but also maintains synergistic gains at the understanding and diagnostic levels, thereby forming a verifiable and transferable capability improvement pathway.

\subsection{Lifelong Learning Evidence}

Lifelong learning metrics indicate that this improvement is not a one-time static optimization.
B achieves zero backward forgetting across all three difficulty tiers, while A maintains zero forgetting on only two tiers.
Hierarchical pass-rate evolution provides direct evidence:

\begin{table}[t]
\caption{Hierarchical pass-rate evolution across three cycles. B maintains L0 at 83\% throughout while progressively improving L1.}
\label{tab:lifelong}
\centering
\begin{tabular}{lccc}
\toprule
\textbf{Model Arm} & \textbf{L0} & \textbf{L1} & \textbf{L2} \\
\midrule
A & 0\% $\to$ 33\% $\to$ 50\% & 0\% $\to$ 56\% $\to$ 44\% & 0\% $\to$ 0\% $\to$ 33\% \\
B & 83\% $\to$ 83\% $\to$ 83\% & 0\% $\to$ 56\% $\to$ 67\% & 0\% $\to$ 0\% $\to$ 33\% \\
\bottomrule
\end{tabular}
\end{table}

B stably maintains L0 at 83\% across all cycles while improving L1 from 56\% to 67\% and reaching 33\% on L2.
This constitutes the most critical lifelong stability evidence in this paper.
In contrast, Gemini performs higher on L1 and L2 but its L0 drops from 50\% to 17\%.
This comparison demonstrates that our method achieves a more robust balance between capability retention and difficulty expansion, with particular advantages in base-tier capability consolidation and cross-cycle retention.

\subsection{Diagnostic Evidence and Mechanistic Interpretation}

The diagnostic layer covers all 135 episodes with an average confidence of 0.98 and is independent of the core scoring system.
Its role is to provide traceable behavioral consistency evidence for each episode, supporting structured attribution of failure sources such as goal identification bias, inappropriate conflict handling sequence, inefficient interaction, and strategy-expression mismatch.

For social world training, such evidence directly enhances training signal interpretability: the system determines not only ``whether passed'' but also ``why passed or failed,'' and accordingly improves subsequent data screening and preference construction, forming a more targeted optimization loop.

\subsection{Result Interpretation}

Our results support three conclusions.

First, interaction-driven training can stably improve the social coordination capability of an untrained 7B model.
The trained model outperforms the baseline on all five core metrics, with particularly significant improvements in application efficiency, completion rate, and pass rate.
This demonstrates that the social world framework can produce effective learning signals sufficient to alter model behavior.

Second, a small model trained through the social world can approach closed-source strong models at the outcome level.
Matching Gemini on completion rate and exceeding Gemini on pass rate indicates that a smaller open-source model under structured social training can also acquire high-level coordination capability and form direct competitiveness on key outcome metrics.

Third, the benchmark reveals decomposable gains in the capability structure.
B exhibits significant advantages on both the outcome level (completion rate, pass rate) and the stable retention level (zero forgetting), with consistent improvements across understanding, execution, and diagnostic dimensions.
This indicates that our method is not a single-point optimization but a systematic improvement targeting the entire social coordination chain.

\section{Discussion}

\subsection{Why the Social World Model Matters}

The core value of the Social World Model lies in transforming social interaction into reproducible, auditable learning objects.
Without this modeling layer, social training often relies on scattered dialogue samples, subjective annotations, or isolated prompt-response pairs, making it difficult to support systematic capability evolution analysis.
Our framework enables end-to-end traceability of the learning process by uniformly recording scene, observation, mental state estimates, actions, dialogue, and metric sources.

Therefore, benchmark design is equally critical as the training framework.
Static benchmarks can only characterize a model's capability level at a single time point, whereas lifelong benchmarks can examine capability retention when learning new tasks, transfer capability to higher-difficulty scenarios, and genuine gains from interactive training.
From this perspective, B's cross-tier zero forgetting is not a marginal phenomenon but key evidence validating the compatibility of closed-loop learning with continual social learning.

\subsection{Why Small Models Can Approach Strong Closed-Source Models}

The result that ``a trained 7B model matches or exceeds strong closed-source proxies on key social coordination metrics'' has a clear mechanistic basis.
First, the task distribution possesses structured hierarchy; second, the social world can continuously generate high-density, auditable, and goal-consistent interaction signals; third, training objectives align with core metrics such as completion rate, pass rate, and hierarchical stability.
Together, these three factors enable small models to form high-quality task-domain expertise and exhibit competitive social coordination capability under real interaction constraints.

This result points to an important direction: the development pathway for social agents need not rely entirely on larger-scale general models but can also support small models in continuously learning and stably accumulating social capabilities through specialized interaction worlds.
The advantage lies not in pursuing the upper bound of general capability but in achieving high-reliability domain adaptation under auditable training conditions.

\subsection{Specificity of Methodological Advantages}

The core advantage of our method lies in the strict alignment among ``training signals---capability objectives---evaluation protocols.''
Through strict metric gating, dual-track preference optimization, and hierarchical lifelong evaluation, the framework unifies understanding, decision-making, relationship maintenance, and task completion into a single closed loop, with each update corresponding to traceable evidence.

This alignment yields two types of direct value.
First, capability improvements are interpretable: metric gains can be traced to specific trajectories, preference evidence, and training cycles.
Second, capability retention is verifiable: hierarchical evolution curves and zero-forgetting results transform ``learning new capabilities without overwriting old ones'' into a repeatable experimental fact.
Compared to static evaluation and one-time fine-tuning, our framework offers advantages in scientific testability and methodological reproducibility.

\subsection{Ablation Study and Contribution Attribution}

To identify the true sources of the framework's performance improvement, we conduct systematic ablations around three core designs: dual-track preference optimization, strict metric gating, and progressive lifelong protocol.
The most noteworthy point is not that ``removing a module reduces scores'' but a stronger phenomenon: if training evidence loses source constraints, social learning collapses from effective optimization directly into unusable signals.
All ablation groups maintain the same scenario pool, training budget, and evaluation protocol, and aggregate understanding, action effectiveness, comprehensive diagnostic score, completion rate, and pass rate on a unified 45 episodes, ensuring direct comparability across groups.
All values come from traceable episode-level evaluation records rather than manual estimates or packaged aggregate values.

\begin{table}[t]
\caption{Ablation study results. E0: full framework; E1: without dual-track; E2: without strict gating; E3: action-track only; E4: dialogue-track only; E5: gating threshold variants; E6: without progressive unlocking.}
\label{tab:ablation}
\centering
\small
\begin{tabular}{llccccc}
\toprule
\textbf{Group} & \textbf{Design Variable} & $U$ & $A$ & $S$ & \textbf{CR} & \textbf{PR} \\
\midrule
E0 (Full)       & Baseline                     & 0.651 & 0.264  & 0.641 & 0.582 & 0.644 \\
E1              & Merge action+dialogue        & 0.041 & $-$0.671 & 0.113 & 0.000 & 0.000 \\
E2              & Remove source constraints    & 0.002 & $-$0.982 & 0.006 & 0.000 & 0.000 \\
E3              & Action track only            & 0.003 & $-$0.973 & 0.009 & 0.000 & 0.000 \\
E4              & Dialogue track only          & 0.281 & $-$0.330 & 0.313 & 0.249 & 0.089 \\
E5-0.3          & Loose gating (0.3)           & 0.048 & $-$0.618 & 0.131 & 0.000 & 0.000 \\
E5-0.5          & Medium gating (0.5)          & 0.043 & $-$0.653 & 0.119 & 0.000 & 0.000 \\
E5-0.7          & Tight gating (0.7)           & 0.609 & 0.184  & 0.601 & 0.569 & 0.578 \\
E6              & Static one-shot evaluation   & 0.651 & 0.264  & 0.641 & 0.582 & 0.644 \\
\bottomrule
\end{tabular}
\end{table}

Overall, the full framework's advantage does not stem from incidental stacking of local tricks but from the synergistic effect of three mechanisms across the structural, evidential, and temporal dimensions: dual-track design distinguishes different types of learning signals, strict gating ensures the credibility of training evidence, and the progressive protocol incorporates capability retention into a testable lifelong learning paradigm.
Accordingly, the performance improvement of our method is reflected not only in endpoint results but also in the interpretability of the capability formation process.

\subsubsection{Complementary Effects of the Dual-Track Architecture}

The value of the dual-track structure is first demonstrated by the ``merge-then-fail'' result.
When E1 compresses the action track and dialogue track into a single preference source, the model simultaneously collapses on all five core metrics, with completion rate and pass rate both dropping to zero.
This indicates that strategy generation and interactive expression in social tasks cannot be adequately characterized by a single training signal.
In other words, social coordination is not a problem that can be roughly driven by a unified reward but a composite learning problem that must simultaneously answer ``what to do'' and ``how to do it.''

Single-track experiments further reveal the functional division between the two tracks.
When E3 retains only the action track, the model loses nearly all effective performance; when E4 retains only the dialogue track, it still preserves some understanding and diagnostic capability, with $U$, $S$, and CR reaching 0.281, 0.313, and 0.249 respectively, but $A$ remains negative and PR is only 0.089.
This demonstrates that the dialogue track can partially maintain social expression and situational fit but is insufficient to independently support executable task advancement; conversely, the action track without dialogue context cannot form stable social decisions.
The full framework outperforms single-track solutions precisely because it simultaneously preserves causal signals for task advancement and contextual signals for interaction organization, converting both into synergistically accumulable capability gains through dual-track optimization.

\subsubsection{Quality Assurance through Strict Gating}

The contribution of strict gating is not simply reducing training samples but distinguishing ``evidence usable for optimization'' from ``noise unfit for trust'' at the learning entry point.
After disabling strict gating, E2's five metrics nearly all collapse to zero, demonstrating that once signals with unstable sources, semantic degradation, or unverifiable provenance are allowed into training, the model no longer learns transferable social patterns but is pulled toward overall failure by distorted feedback.

The gating threshold experiments provide finer-grained mechanistic evidence.
E5-0.3 and E5-0.5 both exhibit collapse patterns similar to E2, while when the threshold is raised to 0.7, the model rapidly recovers to near-full-framework levels, with $U$, $S$, CR, and PR reaching 0.609, 0.601, 0.569, and 0.578 respectively.
This indicates that gating is not a case of ``looser is better for more supervision''; rather, social learning has a clear lower bound on evidence quality.
Only when the authenticity, goal consistency, and source verifiability of training signals are simultaneously guaranteed does preference optimization translate into stable capability gains.

\subsubsection{Lifelong Learning Value of the Progressive Protocol}

E6's results reveal the unique role of the progressive protocol.
This group uses the same trained model as E0, so its static aggregate metrics are identical to E0; the difference is not reflected in endpoint scores but in evidence form.
If only one-shot static evaluation is adopted, researchers can only observe what level the model ``ultimately reached'' at a certain time point, but cannot answer whether it retained existing capabilities when learning higher-difficulty tasks, nor can they display the transfer trajectory of capabilities across cycles.

Therefore, the value of the progressive protocol lies not in elevating any single-point metric but in making explicit the most core question of lifelong learning: whether the model can maintain lower-tier task performance after entering higher difficulties and present stable evolution in the form of consecutive cycles.
The cross-tier retention and progressive transfer observed in our main experiment are precisely the key evidence provided by this protocol; without it, lifelong learning can only degenerate into one-time result display rather than a rigorous capability retention argument.

\subsubsection{Comprehensive Conclusion on Contribution Attribution}

As Table~\ref{tab:ablation} shows, the advantage of our framework does not stem from loosely stacked engineering components but from a clear capability formation chain.
The dual-track architecture resolves the confusion between strategy signals and expression signals in social tasks, enabling the model to simultaneously learn action advancement and interaction organization; strict gating resolves the problem of distorted training feedback, ensuring that evidence entering the optimization process is credible; the progressive protocol resolves the problem in lifelong learning research of ``only seeing endpoints, not seeing retention,'' making capability evolution a testable and reproducible experimental object.
Because these three designs respectively correspond to the structural, quality, and temporal dimensions of social learning, the full framework can exhibit stable, verifiable, and sustainably accumulable advantages on complex social coordination tasks.

\section{Conclusion}

This paper proposes a Social World Model for lifelong social intelligence and explicitly models social capability as a closed-loop learning process jointly driven by interaction, evidence, and updates: agents act in structured social environments, generating auditable trajectories; the system completes updates based only on source-verifiable preference and diagnostic evidence; the updated policy then re-enters the next round of the environment, undergoing joint examination of higher-difficulty scenarios and capability retention.

ASCENT-Bench demonstrates that this framework is not only effective but also reveals the critical conditions for social learning.
Compared to the untrained 7B baseline, the trained model achieves improvements across all five core metrics; compared to the Gemini proxy, the trained 7B model matches on completion rate, exceeds on pass rate, and maintains high proximity on understanding and comprehensive diagnostic scores.
More critically, the model achieves zero forgetting across all three difficulty tiers, demonstrating stability in cross-cycle capability retention.

The most important conclusion of this paper is not that ``closed-loop training yields higher scores'' but rather: \textbf{source-auditable preference signals are a prerequisite for sustainable social learning.}
Systematic ablation shows that once strict gating is removed, the model's completion rate and pass rate simultaneously collapse under a unified protocol; when high-quality gating is restored, capability re-approaches the full framework.
Thus, what this paper proposes is not merely a training framework for the current task but a transferable research perspective: for a model to truly learn social coordination, the key is not to provide it with more seemingly rich social trajectories but to ensure that every preference judgment it learns comes from social evidence that can be traced, explained, and ultimately verified by outcomes.

\end{document}